\documentclass[runningheads]{llncs}
\usepackage{graphicx}
\usepackage{mathtools}
\usepackage{tikz}
\usepackage{comment}
\usepackage{amsmath,amssymb} 
\usepackage{color}
\usepackage{subcaption}

\usepackage[capitalize]{cleveref}
\crefname{section}{Sec.}{Secs.}
\Crefname{section}{Section}{Sections}
\Crefname{table}{Table}{Tables}
\crefname{table}{Tab.}{Tabs.}

\usepackage{tikz}
\usepackage{amsmath}

\newcommand{\RB}{\mathbb{R}}
\newcommand{\x}{\textbf{x}}

\newcommand{\A}{\textbf{A}}
\newcommand{\X}{\textbf{X}}
\newcommand{\Y}{\textbf{Y}}
\usepackage{amssymb}

\usepackage{pifont}

\newcommand{\xmark}{\ding{55}}%

\usepackage[accsupp]{axessibility}  

\begin{document}
\pagestyle{headings}
\mainmatter
\def\ECCVSubNumber{6571}  

\title{Beyond Periodicity: Towards a Unifying Framework for Activations  in Coordinate-MLPs} 


\titlerunning{Beyond Periodicity}
%
\author{Sameera Ramasinghe   \and
Simon Lucey }
\authorrunning{Sameera Ramasinghe and Simon Lucey}
%
\institute{Australian Institute for Machine Learning\\
University of Adelaide\\
\email{sameera.ramasinghe@adelaide.edu.au}}
\maketitle

\begin{abstract}
Coordinate-MLPs are emerging as an effective tool for modeling multidimensional continuous signals, overcoming many drawbacks associated with discrete grid-based approximations. However, coordinate-MLPs with ReLU activations, in their rudimentary form, demonstrate poor performance in representing signals with high fidelity, promoting the need for positional embedding layers. Recently, Sitzmann \textit{et al.}~\cite{sitzmann2020implicit} proposed a sinusoidal activation function that has the capacity to omit positional embedding from coordinate-MLPs while still preserving high signal fidelity. Despite its potential, ReLUs are still dominating the space of coordinate-MLPs; we speculate that this is due to the hyper-sensitivity of networks -- that employ such sinusoidal activations -- to the initialization schemes. In this paper, we attempt to broaden the current understanding of the effect of activations in coordinate-MLPs, and show that there exists a broader class of activations that are suitable for encoding signals. We affirm that sinusoidal activations are only a single example in this class, and propose several \textbf{non-periodic} functions that empirically demonstrate more robust performance against random initializations than sinusoids. Finally, we advocate for a shift towards coordinate-MLPs that employ these non-traditional activation functions due to their high performance and simplicity.
\end{abstract}

\section{Introduction}
\label{sec:intro}

Despite the ubiquitous and successful usage of conventional discrete representations in machine learning (\textit{e.g.} images, 3D meshes, and 3D point clouds etc.), coordinate MLPs are now emerging as a unique instrument that can represent multi-dimensional signals as continuously differentiable entities. Coordinate-MLPs -- also known as \emph{implicit neural representations} \cite{sitzmann2020implicit} -- are fully connected networks that encode continuous signals as weights, consuming low-dimensional coordinates as inputs. Such continuous representations are powerful compared to their discrete grid-based counterparts, as they can be queried up to extremely high resolutions. Furthermore, whereas the memory consumption of grid-based representations entails exponential growth rates against the dimension and the resolution of data, neural representations have displayed a much more compact relationship between the above factors. Consequently, this recent trend has influenced a proliferation of studies in vision-related research including texture generation \cite{henzler2020learning,oechsle2019texture,henzler2020learning,xiang2021neutex}, shape representation \cite{chen2019learning,deng2020nasa,tiwari2021neural,genova2020local,basher2021lightsal,mu2021sdf,park2019deepsdf}, and novel view synthesis \cite{mildenhall2020nerf,niemeyer2020differentiable,saito2019pifu,sitzmann2019scene,yu2021pixelnerf,pumarola2021d,pumarola2021d,rebain2021derf,martin2021nerf,wang2021nerf,park2021nerfies}.

Notwithstanding the virtues mentioned above, coordinate MLPs, in their fundamental form, exhibit poor performance in encoding signals with high-frequency components when equipped with common activation functions such as ReLUs. An elemental reason behind this has shown to be the \emph{spectral bias} of MLPs \cite{basri2020frequency,rahaman2019spectral}. That is, the corresponding neural tangent kernel (NTK) of MLPs are prone to high-frequency fall-offs, hampering their ability to learn high-frequency functions. The prevalent work-around to this problem involves applying a \emph{positional embedding layer} prior to the MLP, where the low-dimensional inputs are projected to a higher-dimensional space using Fourier features \cite{tancik2020fourier}.

In contrast, Sitzmann \textit{et al.}~\cite{sitzmann2020implicit} recently portrayed that MLPs with sinusoidal activation functions are naturally suited for encoding high-frequency signals, eliminating the need for positional embedding layers. Despite its potential, much of the research that involve coordinate-MLPs still prefer positional embeddings over sinusoidal activations. We postulate that this could be for two reasons. First, Sitzmann \textit{et al.} mainly attributed the success of sinusoidal activations to their periodicity, although the evidence for this relationship still remains scant.  Consequently, this lack of understanding obfuscates some of the fundamental principles behind its effectiveness and hampers faithful usage in a wider range of applications. Second, sinusoidal activations are highly sensitive to the initialization of the MLP, showcasing significant performance drops in cases where the MLP is initialized  without strictly adhering to the guidelines of   Sitzmann \textit{et al}. The above drawbacks have heightened the need for a more rigorous analysis that facilitates more effective usage of activation functions in coordinate-MLPs. 

 \noindent{\textbf{Contributions:}} We offer a broader theoretical understanding of the role of activation functions within coordinate-MLPs. In particular, we show that the efficacy of a coordinate-MLP is critically bound to its Lipschitz smoothness and the singular value distribution of the hidden-layer representations, and the optimal values of these metrics depend on the characteristics of the signal that needs to be encoded. We further show that the above properties are inherently linked to each other, and by controlling one property, the other can be implicitly manipulated. We further derive formulae to connect the Lipschitz smoothness and the singular value distribution to the properties of the activation functions. The significance of this finding is two-fold: (i) providing  guidelines for tuning the hyper-parameters of an activation function based on the given signal and, (ii) enabling a practitioner to theoretically predict the effect of a given activation function when used in a coordinate-MLP, prior to practical implementation. We further show that sinusoidal activations are simply a single example that fulfills such constraints, and the periodicity is not a crucial factor that determine the efficacy of an activation function. Consequently, we propose a much broader class of \emph{non-periodic} activation functions that can be used in encoding functions/signals with high fidelity, and show that their empirical properties match with theoretical predictions. We further illustrate that the newly proposed activation functions are robust to different  initialization schemes,  unlike sinusoidal activations. Further, picking one such proposed activation -- Gaussian -- as an example, we demonstrate that coordinate-MLPs with such activation functions enjoy better results, faster convergence rates, and shallower architectures  in comparison to ReLU-MLPs. Finally, we show that these activations allow positional-embedding-free architectures to be used in complex tasks such as 3D view synthesis. To our knowledge, this is the first instance coordinate-MLPs have successfully been employed in such experiments in the absence of positional embeddings.

 \begin{figure}[!tp]
    \centering
    \includegraphics[width=\columnwidth]{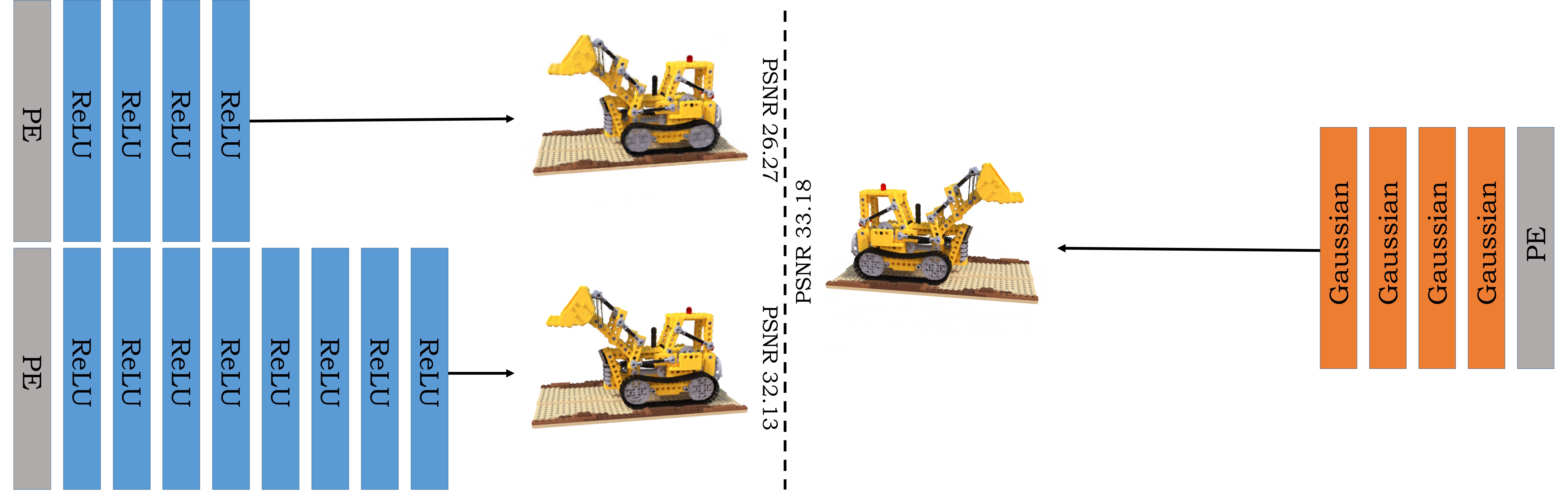}

    \caption{\small \textbf{ReLU vs Gaussian activations (ours).} Gaussian activations achieve better results  with $\sim 50\%$ less parameters. These non-periodic activations also allow embedding-free architectures (see Fig.\ref{fig:reluvsgau}), and are  robust to different random initializations of  coordinate-MLPs than the sinusoid activations advocated in SIREN~\cite{sitzmann2020implicit}.}
    \label{fig:derivatives}
\end{figure}  
 
 \section{Related works}
\noindent{\textbf{Activation functions:}} (Non-linear) activation functions are an essential component of \emph{any} neural network that aims to model the relationship between complex functions and their coordinates. These activations significantly broaden the class of signals/functions one can approximate. The role of activation functions have been studied extensively since the earliest days of neural network research \cite{williams1986logic,baum1988supervised,feldman1988computing,murray1987bit}. Perhaps, the first notable discussion about the theoretical properties of activations was presented by Williams \textit{et al.} \cite{williams1986logic}, where invariance of non-linearities under logical and ordinal preserving transformation of inputs was explored. Dasgupta \textit{et al.} \cite{dasgupta1992power} compared the approximation
power of feedforward neural nets with different activations. This initial work was followed by a series of analytical studies including discontinuous \cite{hopfield1984neurons,hopfield1986computing,lu2005dynamical}, polynomial-based \cite{ma2005constructive} and Lipschitz bounded \cite{liang2000absolute,liang2001additive,cao2004absolute} activation functions. In addition to this, a plethora of empirical research has also  been published \cite{sharma2017activation,sibi2013analysis,ramachandran2017searching,pedamonti2018comparison,karlik2011performance}. Periodic non-linearities have also been explored in terms of Fourier feature networks \cite{gallant1988there,silvescu1999fourier}, recurrent neural networks \cite{liu2015multistability,koplon1997using}, and classification tasks \cite{sopena1999neural,wong2002handwritten}. 

\noindent{\textbf{Coordinate MLPs:}} In recent years, there has been an increasing interest in parameterizing signals using neural networks -- commonly referred to as coordinate-MLPs~\cite{sun2021coil} or implicit neural functions~\cite{sitzmann2020implicit} -- largely owing to the seminal work by Mildenhall \textit{et al.} \cite{mildenhall2020nerf}. The usage of coordinate-MLPs are somewhat different  from conventional MLPs: i) conventional MLPs typically operate on high dimensional inputs such as images, sounds, or 3D shapes, and ii) are primarily being used for classification purposes where the decision boundaries do not have to preserve smoothness. In contrast, coordinate-MLPs are used to encode  the signals as weights where the inputs are low-dimensional coordinates and the outputs have to preserve smoothness \cite{zheng2021rethinking}. One of the most remarkable aspects of  Mildenhall \textit{et al.}'s work includes demonstrating the generalization properties of such neural signal representations, \textit{i.e.} once trained with a handful of view points, the coordinate-MLP can reconstruct the photometric view projection from an arbitrary angle with fine-details. This ground-breaking demonstration caused a ripple of studies that include neural signal representations as the core entities across many applications including shape representation \cite{chen2019learning,deng2020nasa,tiwari2021neural,genova2020local,basher2021lightsal,mu2021sdf,park2019deepsdf}, and novel view synthesis \cite{mildenhall2020nerf,niemeyer2020differentiable,saito2019pifu,sitzmann2019scene,yu2021pixelnerf,pumarola2021d,pumarola2021d,rebain2021derf,martin2021nerf,wang2021nerf,park2021nerfies}. However, for optimal performance, these coordinate-MLPs have to use positional embeddings, which allow them to encode high-frequency signal content. In contrast,  Sitzmann \textit{et al.} \cite{sitzmann2020implicit} proposed sinusoid activations that enabled coordinate-MLPs to encode signals with higher quality without positional embeddings. But, sinusoid activations have been shown to be extremely sensitive to the initialization scheme of the MLPs. A further limitation to the framework developed by  Sitzmann \textit{et al.} is its confinement to periodic activations. In contrast, our work generalizes the current understanding on the effect of activations in coordinate-MLPs and thereby propose a class of non-periodic activations that is robust under random initializations.

\section{Methodology}

\noindent \textbf{Notation.} The set of real $n-$dimensional vectors are denoted by $\RB^n$. The vectors are denoted by  bold lower-case letters (\textit{e.g.,} $\x$). The set of $m \times n$ dimensional matrices are denoted by $\RB^{m \times n}$, and the matrices are denoted by bold upper-case letters (\textit{e.g.,} $\A$).  $\| \cdot \|$ denotes vector norm,  $\| \cdot \|_F$ denotes the Frobenius norm, and $\| \cdot \|_o$ is the operator norm. $\mathbb{B}^n_r$ represents  the $n-$ dimensional ball with radius $r$. Further, $g(f(\x)) = g \circ f(x)$ where $\circ$ is the compositional operator.


\subsection{Rank and memorization}
\label{sec:rank}

The efficacy of a coordinate-MLP  largely depends on its ability to memorize training data. The objective of this section is to identify the key factors that affect memorization. To establish the foundation for our analysis, we first revisit the formulation of an MLP.

An MLP $f$ with $k-1$ non linear hidden-layers can be described by,

\begin{equation}
    f: \x \to g^k \circ \psi^{k-1} \circ g^{k-1} \circ \dots \circ \psi^1 \circ g^1(\x),
\end{equation}

where $g^i: \x \to \A^i \cdot \x + \textbf{b}^i$ is an affine projection with trainable weights $\A^i \in \RB^{\mathrm{dim}(\x^{i}) \times \mathrm{dim}(\x^{i-1})}$, $\textbf{b}^i \in \RB^{\mathrm{dim}(\x_{i})}$ is the bias,  and $\psi^i$ is a non-linear activation function.  The final layer is a linear transform such that $f: \x \to g^{k} \circ \phi(\x)$, and $\phi$ is a composition of the preceding $k-1$ layers within the MLP without the final linear layer. If the number of training examples is $N$, we define the total (training) embedding matrix as
\begin{equation}
    \X \in \RB^{D \times N}  \coloneqq \begin{bmatrix}
       \phi(\x_1)^T \phi(\x_2)^T \dots  \phi(\x_N)^T\\
    \end{bmatrix}
\end{equation}
where~$\{\x_{n}\}_{n=1}^{N}$ are the raw training inputs. 

Recall that the final layer of an MLP is (typically) an affine projection without any non-linearity. Dropping the bias for simplified notation, we get,

\begin{equation}
    \tilde{\Y} = \A^k\X ,
    \label{Eq:equals}
\end{equation}
where $\tilde{\Y} \in \RB^{q \times N}$ are the outputs of the MLP. Suppose $\Y \in \RB^{q \times N}$ are the ground truth training outputs the MLP is attempting to learn. Observe that if the MLP is perfectly memorizing the training set --- if $\tilde{\Y} = \Y$ --- then
 each row of $\Y$ is a linear combination of the rows of $\X$. Assume we have no prior knowledge of $\Y$, that is, the rows of $\Y$ can be \emph{any} arbitrary vector in $\RB^N$. If the rows of $\X$ are linearly independent, they form a basis for $\RB^N$ (assuming $D \geq N)$. Therefore, if $\mathrm{rank}(\X) = N$, it is guaranteed that (assuming perfect convergence) the MLP  can find a weight matrix $\A^k$ that ensures perfect reconstruction of $\Y$.   
 
 One can raise the valid question:  could this conclusion hold in the practical case where $D \ll N$? The answer to this question depends on the nature of the ground truth signal. Note that although the condition $\mathrm{rank}(\textbf{X}) = N$ is sufficient to ensure perfect memorization for \emph{any} signal, it might not always be necessary since natural signals are typically redundant -- that is of limited bandwidth. The bandwidth of a category of signals can be defined~\cite{shannon1949communication} as the number of linearly independent  (normalized) bases required to represent them. Thus, $\mathrm{rank}(\textbf{X})$ can be less than $N$ for many categories of signals whilst still enjoying perfect signal recovery by the MLP. Fig. \ref{fig:activations} is a perfect example that illustrates the above point. Note that the stable rank is a lower bound for rank \cite{rudelson2007sampling}. Better reconstructions are shown when the stable rank is high, but the measure is bounded by the network width ($D$), which is lower than the number of points ($N$).  In contrast,  encoding noise signals which have limited to no redundancy -- would require a larger network width -- and yields poorer results with $D \ll N$ (Table \ref{tab:2dapp}, Appendix) as predicted. Rigorously speaking, the analysis so far only considers the penultimate layer. However, based on the gathered insights, we make the following general claim: \emph{the potential of the hidden-layers to induce high-rank representations -- that is those with very few zero singular values within~$\X$ -- correlates with the memorization capacity of a coordinate-MLP.} 
 
 One could also view the above result as a refashioning of the well known Nyquist-Shannon sampling theory~\cite{shannon1949communication} applied to-coordinate MLPs. The result is important, however, when it comes to the exposition of the rest of this paper. But, a critical component is overlooked in the above analysis. In many applications that utilize coordinate-MLPs, the ability predict values at unseen coordinates, \textit{i.e.}, generalization, is important. For instance, in novel view synthesis of a 3D scene, the network only observes a handful of views, in which the network then has to predict the views from new angles. Therefore, the immediate question arises: \emph{is having the ability to induce high-rank representations (i.e. very few zero singular values within~$\X$) sufficient  for both memorization and generalization?} In Section \ref{sec:Smoothness}, we shall see that this is indeed not the case.

\subsection{Smoothness and generalization}
\label{sec:Smoothness}

To show that the rank alone is not sufficient to guarantee good generalization, we perform a simple thought experiment on 1-D input coordinates~$x \in \mathbb{R}$ and single channel outputs. Let us construct a very wide layer $\phi: \mathbb{R}\to \mathbb{R}^D$ such that $D = N$, and define the layer output  $\phi(x)=[e^{\frac{-(x-x_1)^2}{\sigma^2}}, \dots, e^{\frac{-(x-x_N)^2}{\sigma^2}}]$, where $x_1,\dots, x_N$ are the training points. With extremely small $\sigma$, $\phi(\cdot)$ is equivalent to one-hot encoding, ensuring  $\mathrm{rank}(\textbf{X}) = N$. Then, it is guaranteed that an $\textbf{A}$ can be found to memorize all the ground truth outputs $y_1, \dots, y_N$. However, all the unseen points will map to $\vec{\textbf{0}}$, and thus, the network will be obtain extremely poor generalization. In summary, having a higher rank for $\textbf{X}$ will help in memorization, but, it will not necessarily ensure good generalization. 

Moreover, strictly speaking, generalization cannot be measured independently without context. For instance, given sparse training points a neural network can, in theory, learn infinitely many functions while fitting the training points. Therefore, for good generalization, the network has to learn a function within a space restricted by certain priors and intuitions regarding the ground truth signal. The generalization then depends on the extent to which the learned function is close to these prior assumptions about the task. When no such priors are available, one intuitive solution that is widely accepted for regression (at least from an engineering perspective) is to have ``smooth" interpolation between the training points  \cite{biship2007pattern}. 

 


In order to ensure such smooth interpolations (where second order derivatives are bounded) it is critical to preserve the smoothness across non-linear layers $\phi(\cdot)$ \emph{locally} as $\frac{||\phi(\x_1) - \phi(\x_2)||}{||\x_1 - \x_2||} = C$, where C is some constant (since the final layer is linear). Although the above condition seems overly restrictive, recall that the embeddings $\phi(\cdot)$ are learned via hidden-layers, as opposed to being analytically designed. Therefore, it is enough to reduce the search space of the parameters accordingly, as opposed to explicitly enforcing the above constraint. Thus, we can slightly relax the above equality to an inequality interms of the local Lipschitz smoothness. More precisely, in practice, it is enough to ensure

\begin{equation}
    \| \phi(\x_1) - \phi(\x_2) \| \leq C \| \x_1- \x_2 \|,
\end{equation}

locally, where $C$ is a non-negative,  locally varying constant that depends on the magnitude of the local first-order derivatives (\textit{i.e.}, frequencies) of the encoded signal. That is, at intervals where encoding  points exhibit  large fluctuations, $C$ needs to be higher, and vise-versa.


Thus far, we have established that the singular values of the $\X$ correlates with the memorization of seen coordinates and the (Lipschitz) smoothness of $\phi(\cdot)$ correlates with the generalization performance of an MLP. Thus, it is intriguing to investigate if there exists a connection between these two forces at a fundamental level, as such an analysis has the potential to provide valuable insights that enable efficient manipulation of these factors. 

\subsection{Singular value distribution as a proxy for smoothness}
\label{sec:smoothvsrank}

This section is devoted to exploring the interrelation between the smoothness and the singular value distribution of the hidden representations. Suppose that for coordinates $\x_i$ in a given small neighborhood, $\phi(\cdot)$ is Lipschitz bounded with a constant $C$. Then,

\begin{equation}
\label{equ:lipschitz_angle}
  \frac{\sqrt{  ( \phi(\x_1)\phi(\x_1)^T - 2\phi(\x_1)\phi(\x_2)^T + \phi(\x_2)\phi(\x_2)^T)}}{\| \x_1- \x_2 \|} \leq C 
\end{equation}


\noindent With Eq.~\ref{equ:lipschitz_angle} in hand, let us consider two cases for $\X$.

\begin{case}
The columns of $\X$ are orthogonal and the singular values of $\X$ are identically distributed.
\end{case}


\noindent One can see that,

\begin{equation}
\frac{\sqrt{\|\phi(\textbf{x}_i)\|^2 + \|\phi(\textbf{x}_j)\|^2 }}{\|\textbf{x}_i - \textbf{x}_j\|} \leq C \Rightarrow
     \underset{\| \x_1- \x_2 \|\to 0}{\lim}C = \infty
\end{equation}

 In other words, having (approximately) equally distributed singular values violates the Lipschitz smoothness of the network. 

\begin{case}
The singular values of $\X$ are non-zero and the angle between the columns of $\X$ are upper-bounded by $0 < \alpha < \frac{\pi}{2}$. 
\end{case}

\noindent Consider 

\[
    \textbf{x}^*_i, \textbf{x}^*_j = \underset{\textbf{x}_i,\textbf{x}_i}{\mathrm{arg}}\Big( \frac{\|\phi(\textbf{x}_i) - \phi(\textbf{x}_j) \|}{\|\textbf{x}_i - \textbf{x}_j\|} = C   \Big).
\]

\noindent Then, we can define an upper bound on $C$ as

\[
   C \leq \frac{\sqrt{\|\phi(\textbf{x}^*_i)\|^2 + \|\phi(\textbf{x}^*_j)\|^2 - 2\|\phi(\textbf{x}^*_i)\|\|\phi(\textbf{x}^*_j)\|\mathrm{cos}\alpha}}{\|\textbf{x}^*_i - \textbf{x}^*_j\|},
\]



which can be minimized by decreasing $\alpha$.  Strictly speaking, $C$ can still be considerable with a small $\alpha$, if $|(\|\phi(\textbf{x}^*_i)\| - \|\phi(\textbf{x}^*_j)\|)|$ is large enough. However, in practice, we observe that the $\|\phi(\x)\|$'s do  not deviate from their maximum norm  within the set significantly. That is, within a small sub set of $\x$, the vectors $\phi(\x)$ approximately lie on a sphere (see Appendix).  Therefore, we make the following claim: \textit{the local Lipschitz constant of a network layer can be minimized by reducing the angles between the output vectors}. Below, we justify this claim from another perspective.

\begin{figure*}[!tp]
    \centering
    \includegraphics[width=\columnwidth]{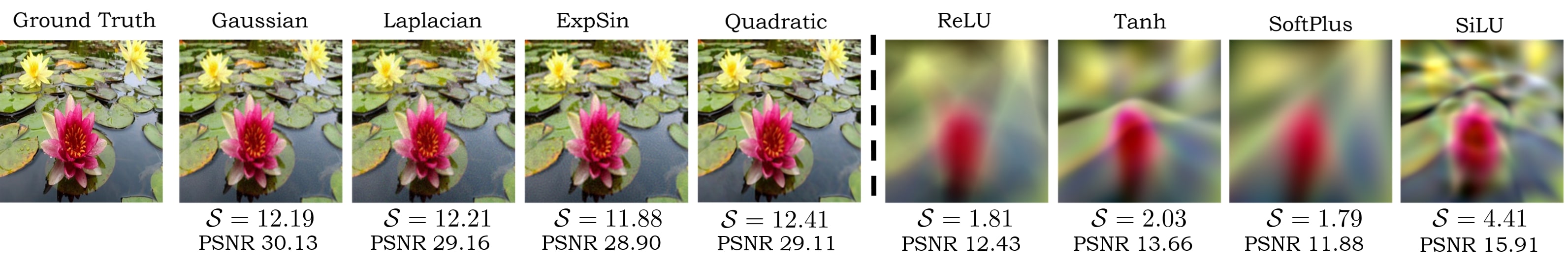}

    \caption{\small \textbf{Proposed activations (left block) vs. existing activations (right block) and their respective stable ranks ($\mathcal{S}$) in image encoding without positional embeddings.} As predicted by Table \ref{tab:activations}, the proposed activations are better suited for encoding signals with high fidelity. As Sec.~\ref{sec:Smoothness} stated, the stable ranks of the proposed activations are higher, indicating larger local Lipschitz constants which allow sharper edges.}
    \label{fig:activations}
\end{figure*}

Consider a set of coordinates $\{\x_i\}_{i=1}^N$ and the function $\phi(\cdot)$ induced by a hidden-layer of an MLP. Let $\{\lambda\}_{i=1}^N$ be the singular values of $\X$ where $\X = \begin{bmatrix}
       \phi(\x_1)^T \phi(\x_2)^T \dots  \phi(\x_N)^T\\
    \end{bmatrix}$. Intuitively, if the angles $\alpha$ between the columns of $\textbf{X}$ are small, most of the energy of the singular values should be concentrated on the first few components. On the other hand, if $\alpha$ is high, the energies should be  distributed. Therefore, we advocate in this paper that the stable rank, defined as $\mathcal{S}(\X) =  \sum_{i=1}^N \frac{\sqrt{\lambda_i}}{\underset{i}{\mathrm{max}}(\sqrt{\lambda_i)}}$ \cite{rudelson2007sampling}, can be used as a useful proxy measure for the spread (angles) of the column vectors of $\textbf{X}$, \textit{i.e.}, $\mathcal{S}(\X)$ is large if the spread is large, and vise-versa. We empirically demonstrate that MLPs are \emph{not} able to obtain a high Lipschitz constant with a small $\mathcal{S}(\X)$ (see Fig.~\ref{fig:activations} and Fig. \ref{fig:params}). If our intuition was incorrect (\textit{i.e.}, if the network could obtain a high Lipschitz constant with a small $\alpha$ by varying the norm of the layer outputs significantly), we should be able to observe high local Lipschitz constants with small $\mathcal{S}(\X)$. Our experimental results in  strongly counters this. That is, networks can \emph{not} obtain a high Lipschitz constant if $\mathcal{S}(\X)$ are low. In other words, coordinate-MLPs primarily try to increase the Lipschitz constant by increasing the angles between the network outputs. 
    
    Based on the gathered insights within this section, we argue that $\mathcal{S}(\X)$ is a potentially useful proxy measure for the local Lipschitz smoothness of network layers. More precisely, if $\mathcal{S}(\X)$ is larger, then the Lipschitz constant $C$ tends to become larger, and vice-versa. This is a useful result, as computing the exact Lipschitz constant of an MLP is an NP-hard problem \cite{scaman2018lipschitz}. Although one can efficiently obtain upper-bounds for the Lipschitz constant, that requires calculating the gradients of the function. Instead, we can gain a rough understanding on the behavior of the Lipschitz smoothness of a particular layer by observing $\mathcal{S}$ at run-time. We should emphasize that these insights are based on intuition and empirical evaluation. A more rigorous proof on this relationship is outside the scope of this paper, but the established relationship is sufficient to allow us to make some useful architectural predictions. In Section \ref{sec:locallipschitz}, we will connect these gained insights to the \emph{local} Lipshchitz smoothness of the signal and the properties of activation functions.

\subsection{Local Lipschitz smoothness and the activation function}
\label{sec:locallipschitz}

\begin{center}
\begin{table*}
\centering
 \resizebox{\columnwidth}{!}{
\begin{tabular}{||c|c|c|c|c|c|c||} 
 \hline
 Activation ($\psi$) & Equation & parameterized & $\psi'$ & $\psi''$ & \textbf{R1} &  \textbf{R2}  \\ [0.5ex] 
 \hline\hline
 ReLU & $\mathrm{max}(0,x)$& \xmark & $\begin{cases}
    1, & \text{if } x > 0 \\
    0,              & \text{otherwise}
\end{cases}$ & $0$ &  \xmark & \xmark\\
PReLU &   $\begin{cases}
    x, & \text{if } x > 0 \\
    ax,              & \text{otherwise}
\end{cases}$ & \checkmark & $\begin{cases}
    1, & \text{if } x > 0 \\
    a,              & \text{otherwise}
\end{cases}$ &  $0$ & \checkmark & \xmark \\
 Sin & $\mathrm{sin}(ax)$ & \checkmark & $a\mathrm{cos}(ax)$ & $-a^2\mathrm{sin}(ax)$ & \checkmark & \checkmark\\
 Tanh & $\frac{e^x - e^{-x}}{e^x + e^{-x}}$ & \xmark & $\frac{4\mathrm{e}^{2x}}{\left(\mathrm{e}^{2x}+1\right)^2}$ & $-\frac{8\left(\mathrm{e}^{2x}-1\right)\mathrm{e}^{2x}}{\left(\mathrm{e}^{2x}+1\right)^3}$ & \xmark & \checkmark\\
 Sigmoid & $\frac{1}{1+e^{-x}}$ & \xmark & $\frac{\mathrm{e}^x}{\left(\mathrm{e}^x+1\right)^2}$ & $-\frac{\left(\mathrm{e}^x-1\right)\mathrm{e}^x}{\left(\mathrm{e}^x+1\right)^3}$ & \xmark & \checkmark \\
 SiLU & $\frac{x}{1+e^{-x}}$ & \xmark & $\frac{\mathrm{e}^x\left(\mathrm{e}^x+x+1\right)}{\left(\mathrm{e}^x+1\right)^2}$ & $-\frac{\mathrm{e}^x\left(\left(x-2\right)\mathrm{e}^x-x-2\right)}{\left(\mathrm{e}^x+1\right)^3}$ & \xmark & \checkmark \\
 SoftPlus & $\frac{1}{a}\mathrm{log}(1 + e^{ax})$ & \checkmark & $\frac{e^{cx}}{1 + e^{cx}}$ & $\frac{c\mathrm{e}^{cx}}{\left(\mathrm{e}^{cx}+1\right)^2}$ & \checkmark & \xmark\\
 \hline
 \hline
 Gaussian & $e^\frac{-0.5x^2}{a^2}$ & \checkmark & $-\frac{x\mathrm{e}^{-\frac{x^2}{2a^2}}}{a^2}$ & $\frac{\left(x^2-a^2\right)\mathrm{e}^{-\frac{x^2}{2a^2}}}{a^4}$ & \checkmark & \checkmark\\
Quadratic & $\frac{1}{1 + (ax)^2}$ & \checkmark & $-\frac{2a^2x}{\left(a^2x^2+1\right)^2}$ & $\frac{2a^2\left(3a^2x^2-1\right)}{\left(a^2x^2+1\right)^3}$ & \checkmark & \checkmark\\
Multi Quadratic & $\frac{1}{\sqrt{1 + (ax)^2}}$ & \checkmark & $-\frac{a^2x}{\left(a^2x^2+1\right)^\frac{3}{2}}$ & $\frac{2a^4x^2-a^2}{\left(a^2x^2+1\right)^\frac{5}{2}}$ & \checkmark & \checkmark\\
Laplacian & $e^(\frac{-|x|}{a})$ & \checkmark  & $\frac{x\mathrm{e}^\frac{\left|x\right|}{a}}{a\left|x\right|}$ & $\frac{\mathrm{e}^\frac{\left|x\right|}{a}}{a^2}$   & \checkmark & \checkmark\\
Super-Gaussian & $[e^\frac{-0.5x^2}{a^2}]^{b}$ & \checkmark & $-\frac{bx\mathrm{e}^{-\frac{bx^2}{2a^2}}}{a^2}$ & $\frac{b\left(bx^2-a^2\right)\mathrm{e}^{-\frac{bx^2}{2a^2}}}{a^4}$ & \checkmark & \checkmark\\
ExpSin & $e^{-\mathrm{sin}(ax)}$ & \checkmark & $a\mathrm{e}^{\sin\left(ax\right)}\cos\left(ax\right)$ &$-a^2\mathrm{e}^{\sin\left(ax\right)}\left(\sin\left(ax\right)-\cos^2\left(ax\right)\right)$ & \checkmark & \checkmark\\
\hline
\end{tabular}}
\caption{\textbf{Comparison of existing activation functions (top block) against the proposed activation functions (bottom block).} The proposed activations and the sine activations fulfill \textbf{R1} and \textbf{R2}, implying better suitability to encode high-frequency signals.}
\label{tab:activations}
\end{table*}
\end{center}
Natural signals have varying local Lipschitz smoothness. For instance, an image may contain high variations within a particular subset of the pixels and may consist of constant pixel values within another subset. Since the final layer of an MLP is linear, the hidden non-linear layers should then have the ability to construct representations with varying local Lipschitz smoothness for better signal encoding. In this section, we show that this ability is primarily linked to the first and second-order gradients of the activation function.   

 In Sec.~\ref{sec:smoothvsrank}, we established that in practice, the angle between the network outputs determines the Lipschitz smoothness. It is easy to see that both the affine transformation and the activation function contribute to the composite Lipschitz constant of a hidden layer. However, the Lipschitz constant of the affine transformation is the operator norm of its weight matrix: Let $\x \in \mathbb{B}^m_{\delta}$ with center $\x_0$. Then as $\underset{\delta \to 0}{\lim}$,

\begin{equation}
    \| (\A \x + \textbf{b}) -  (\A \x_0 + \textbf{b})\| \leq C_{\x_0, \delta} \|\x - \x_0 \|
\end{equation}
\begin{equation}
   C_{\x_0, \delta}  =   \underset{\|\x - \x_0 \| \neq 0 }{\sup} \frac{ \| \A (\x-   \x_0)\|}{\|\x - \x_0 \|},
\end{equation}

which is not a local property. In other words, the network can only control the Lipschitz smoothness of the network via the affine layer globally, which is not useful in encoding natural signals. Hence, we direct our attention towards the activation function. However, it is not trivial to establish the connection between the point-wise activation function $\psi: \mathbb{R} \to \mathbb{R}$ and the composite Lipschitz smoothness, given that the vector norms stays approximately the same (which is our empirical observation). Hence, we strive to obtain mathematical  intuition as described next.

Consider an input vector $\x_0 = [x_1, \dots, x_N]$. Further, let $\x_{\epsilon_1} = [x_1 + \epsilon_1, \dots, x_N + \epsilon_1]$ and  $\x_{\epsilon_2} = [x_1 + \epsilon_2, \dots, x_N + \epsilon_2]$. Our intention is to obtain a measure for $\measuredangle(\x_0, \x_{\epsilon_1})   - \measuredangle(\x_0, \x_{\epsilon_2})$. Further,

\[
    \measuredangle(\x_0, \x_{\epsilon_1}) = \mathrm{cos}^{-1} \Big( \frac{\psi(\x_0) \cdot \psi(\x_{\epsilon_1})}{\|\psi(\x_0) \| \|\psi(\x_{\epsilon_1})\|} \Big).
\]

Since the norms are approximately constant, we can use a proxy for $\measuredangle(\x_0, \x_{\epsilon_1})   - \measuredangle(\x_0, \x_{\epsilon_2})$ as,

\begin{align*}
|\tilde{\measuredangle}(\x_0, \x_{\epsilon_1})   - \tilde{\measuredangle}(\x_0, \x_{\epsilon_2})| & = |\psi(\x_0) \cdot \psi(\x_{\epsilon_1}) - \psi(\x_0) \cdot \psi(\x_{\epsilon_2})| \\
& = |\sum_{i=i}^{N}\Big( \psi(x_i + \epsilon_1) - \psi(x_i + \epsilon_2) \Big)\psi(x_i)| \\
& \leq \sum_{i=i}^{N}|\Big( \psi(x_i + \epsilon_1) - \psi(x_i + \epsilon_2) \Big)||\psi(x_i)|\\
& \leq C_{\psi}|\epsilon_1 - \epsilon_2| \sum_{i=i}^{N} |\psi(x_i)|,
\end{align*}

\begin{figure*}[!htp]
    \centering
    \includegraphics[width=\columnwidth]{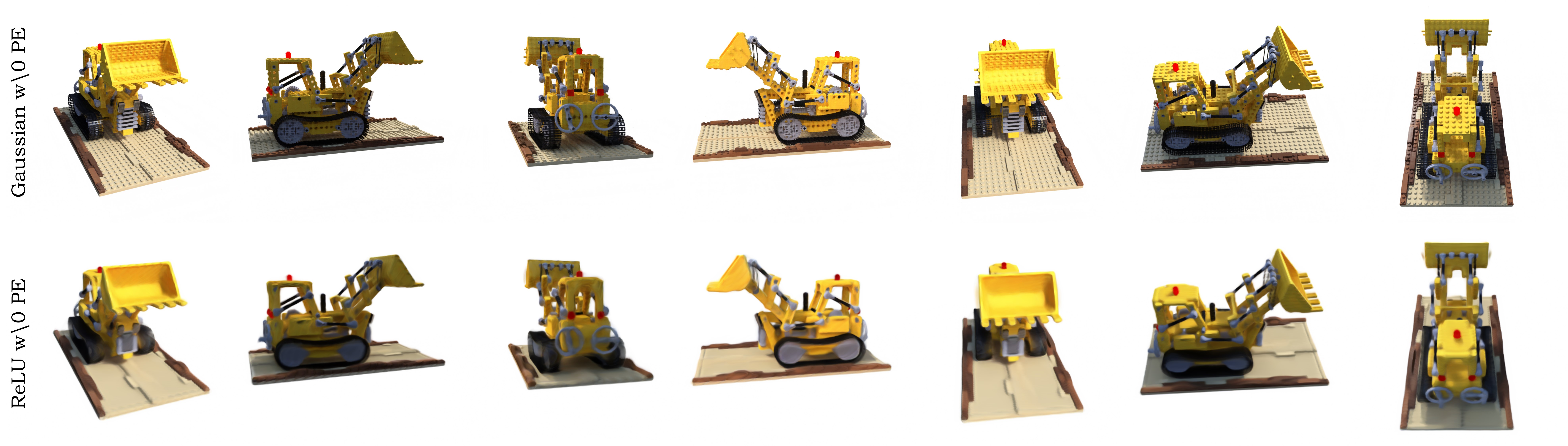}

    \caption{\small \textbf{Novel view synthesis without positional embedding (zoom in for a better view).}  Gaussian activations can completely omit positional embeddings while producing results with significantly better fidelity. In contrast, the performance of ReLU-MLPs severely degrade when positional embeddings are not used. We use 8-Layer MLPs for this experiment.}
    \label{fig:reluvsgau}
\end{figure*}

where $C_\psi$ is the local Lipschits constant of the activation function in the corresponding interval $I$. We then obtain, 

\begin{align}
\frac{|\tilde{\measuredangle}(\x_0, \x_{\epsilon_1})   - \tilde{\measuredangle}(\x_0, \x_{\epsilon_2})|}{|\epsilon_1 - \epsilon_2|} \leq C_{\psi}| \sum_{i=i}^{N} |\psi(x_i)|
\end{align}
Therefore, the upper-bound on the Lipschitz constant of the angle variation in a local interval can be increased by increasing the local Lipschitz constant of the activation function. Further, by definition, the local Lipschits constant  $C_{\psi} = \underset{x \in I}{\sup} (|\frac{d\psi}{dx}|)$. Therefore, we come to the conclusion that in order to encode signals with high frequencies (large fluctuations), one needs to use activation functions that contain first-order derivatives with large magnitudes (the converse is also true). Also, it is important to note that the magnitudes of the local variations depend on the signal. For instance, one can have an extremely smooth signal which can be encoded using activation functions with smaller magnitudes of first-order derivatives. However, the same activations would not be suitable for encoding signals with large fluctuations. Therefore, for better usability across signals with different smoothness properties, activation functions need to be parameterized where the first-order derivates can be controlled via the hyperparameters. We denote this as the requirement 1 (\textbf{R1}) 

However, \textbf{R1} is not necessarily sufficient to ensure good signal fidelity when considering a particular signal with significantly varying fluctuations across different intervals. Thus, for better performance, activations should consist of varying first-order derivatives across a considerable interval, and equivalently, non-negligible second-order derivatives (to obtain varying Lipschitz smoothness). We denote this as requirement 2 (\textbf{R2}). This gives the affine transformations the ability to project the points to different regions of the activation function and achieve varying local Lipschitz smoothness.

\begin{figure*}[!htp]
    \centering
    \includegraphics[width=\columnwidth]{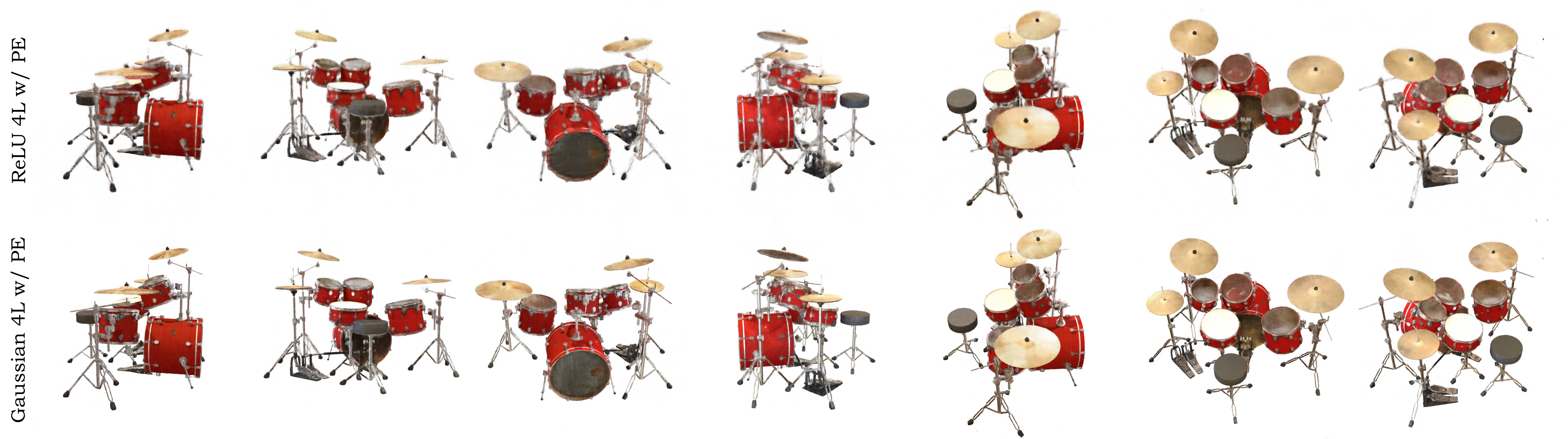}

    \caption{\small \textbf{Novel view synthesis with positional embedding (zoom in for a better view).} With Gaussian activations, shallow MLPs can obtain high-fidelity reconstructions. In contrast, the performance of ReLU-MLPs degrade when the depth of the MLP reduces. We use 4-layer MLPs for this comparison.}
    \label{fig:reluvsgau2}
\end{figure*} 

It is interesting to note that most of the commonly used activations in deep learning do not satisfy above  properties. For instance, consider the ReLU activation $\psi(x) = \mathrm{max}(0,x)$. The derivative of the ReLU then cannot be more than $1$, which hampers its ability to encode function with large local variations. Activations such as Sigmoid, Tanh, GELU also have bounded first-order gradient magnitudes within a smaller range and are not parameterized, which violates \textbf{R1}.  On the other hand, PReLUs, $\psi(x) = \begin{cases}
    x, & \text{if } x > 0 \\
    ax,              & \text{otherwise}
\end{cases}$,  is a parameterized activation that can have extremely large derivatives by controlling the hyperparameter $a$. However, this derivative is either $1$ or $a$, which violates \textbf{R2} and hampers the network's ability to obtain varying local smoothness. In contrast,  recently proposed sine activations \cite{sitzmann2020implicit} $\psi = \mathrm{sin}(ax)$ satisfy both \textbf{R1} and \textbf{R2}, and thus, are suitable for encoding signals. However, we show that the periodicity, as advocated in \cite{sitzmann2020implicit}, is not a crucial requirement, as long as \textbf{R1} and \textbf{R2} are satisfied. Instead, we affirm that there is a much broader class of activations that can be used in coordinate-MLPs, and propose several parameterized activation functions that originate from the family of  infinitely differentiable functions as examples. Table \ref{tab:activations} compares existing and several novel activation functions that we propose, against \textbf{R1} and \textbf{R2}. Finally, it is important to note that even without the restriction that the norms of the vectors are approximately constant, the above conclusions hold (see Appendix).

\begin{figure}[!htp]
\centering
\begin{minipage}[b]{\dimexpr 0.5\textwidth-0.4\columnsep}
\scriptsize
\centering
   \begin{tabular}{||c|c|c|c|c||}
\hline
Activation & Depth & PE & PSNR & SSIM \\
\hline
\hline
ReLU & 4L & \checkmark & 27.44 & 0.922 \\
Gaussian & 4L & \checkmark & \textbf{31.13 } &  \textbf{0.947}\\
\hline
\hline
ReLU & 8L & \xmark & 26.55 & 0.918 \\
Gaussian & 8L & \xmark & \textbf{31.17 }& \textbf{0.949}\\
\hline
\hline
ReLU & 8L & \checkmark & 30.91 & 0.941 \\
Gaussian & 8L & \checkmark & \textbf{31.58}  & \textbf{0.951} \\
\hline
\end{tabular}

  \caption{ \textbf{Quantitative comparison in novel view synthesis on the real synthetic dataset \cite{mildenhall2020nerf}.}  Gaussian activations can achieve high-fidelity reconstructions without positional embeddings. When equipped with positional embeddings, they demonstrate similar performance with $\sim 50\%$ less parameters.}
\label{tab:3D_quantitative}
\end{minipage}%
 \hfill
\begin{minipage}[b]{\dimexpr 0.5\textwidth-0.4\columnsep}
\centering
 \includegraphics[width=\columnwidth]{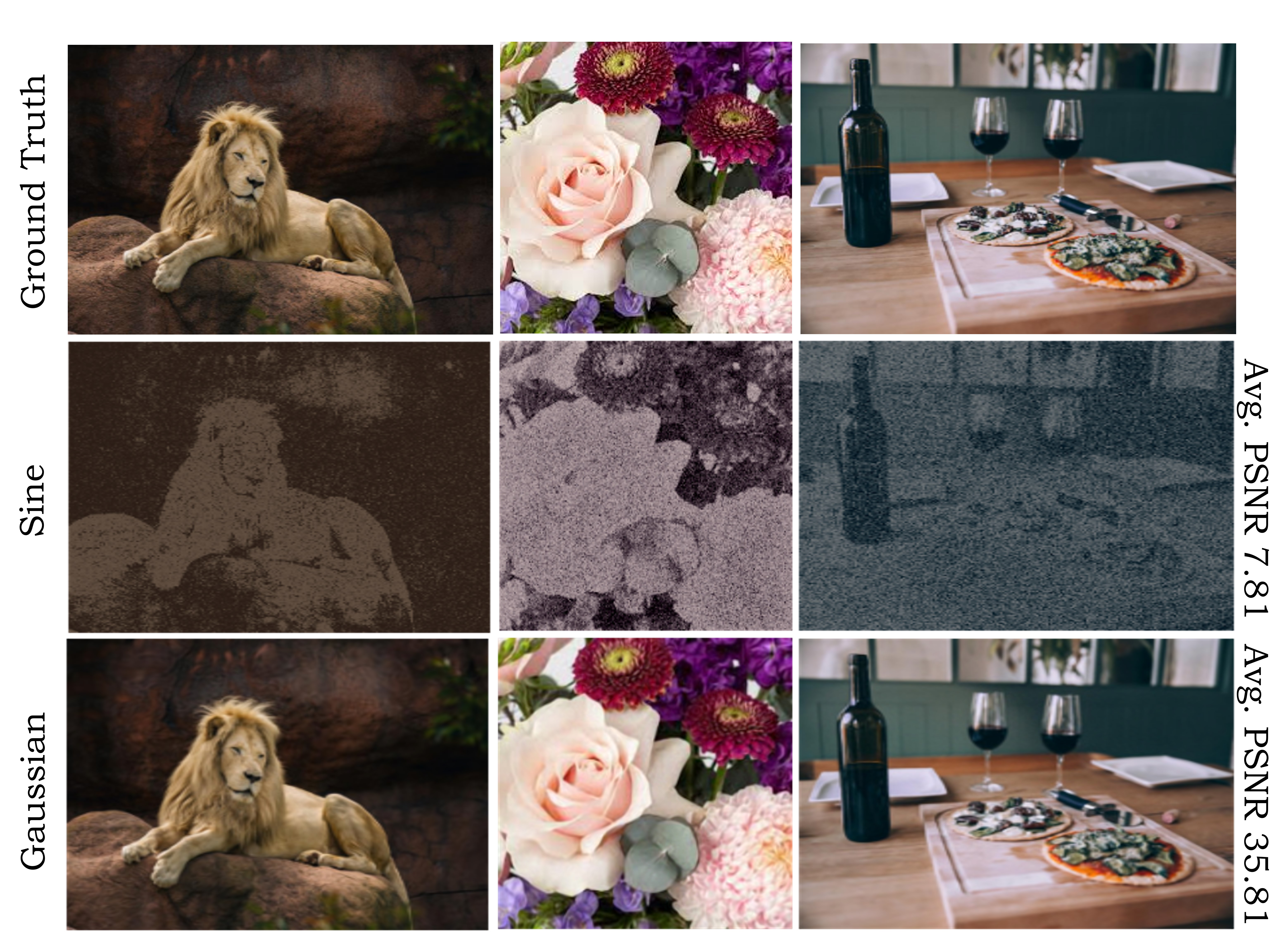}

    \caption{ \textbf{Qualitative comparison of convergence when MLPs are  initialized without following \cite{sitzmann2020implicit} (after 3000 epochs).} Unlike sine activations, Gaussian activations are robust to various initialization schemes (example shown used Xavier normal initialization).}
    \label{fig:convergence}
\end{minipage}
\end{figure}


\section{Experiments}

In this section, we empirically validate the insights we gathered so far.

\subsection{Comparison of activation functions}

We compare the capacity of a coordinate-MLP in encoding signals when equipped with different activation functions. Fig.~\ref{fig:activations} illustrates an example where an image is encoded as the weights of an MLP. As shown, newly proposed Gaussian, Laplacian, ExpSin, and Quadratic activation functions are able to encode the image with significantly better fidelity with sharper gradients (high Lipschitz constants), compared to the existing activations such as ReLU, Tanh, SoftPlus, and SiLU. Also, note that the stable ranks (the energy distribution between the singular values) of the hidden representations are higher for the proposed activation functions than the rest. This matches with our theoretical predictions in Sec.\ref{sec:Smoothness}

\subsection{Novel view synthesis}

\noindent{\textbf{Without positional embeddings:}} We leverage the real synthetic dataset released by \cite{mildenhall2020nerf} to test the capacity of the Gaussian activations in encoding high-dimensional signals. Fig.~\ref{fig:reluvsgau} qualitatively contrasts the performance of ReLU vs. Gaussian activations without the positional embeddings. When the positional embeddings are not used, the ReLU MLPs demonstrate poor performance in capturing high-frequency details. On the contrary, Gaussian activations can capture information with higher fidelity in the absence of positional embedding. We believe this is an interesting result that opens up the possibility of positional-embedding-free architectures.

      

\noindent{\textbf{With positional embeddings:}} Although suitably chosen activation functions can omit positional embeddings, the combination of the two can  still  enable shallower networks to learn high-frequency functions. Fig.~\ref{fig:reluvsgau2} depicts an example with 4-layer MLPs. As evident, when the network is shallower, ReLU MLPs showcase reduced quality, while the performance of Gaussian activated MLPs is on-par with deeper ReLU MLPs. This advocates that practitioners can enjoy significantly cheaper architectures when properly designed activation functions are used. Table \ref{tab:3D_quantitative} depicts the quantitative results that include above comparisons.

\begin{figure}[!htp]
\centering
\begin{minipage}[t]{\dimexpr 0.5\textwidth-0.4\columnsep}
\centering
    \includegraphics[width=\columnwidth]{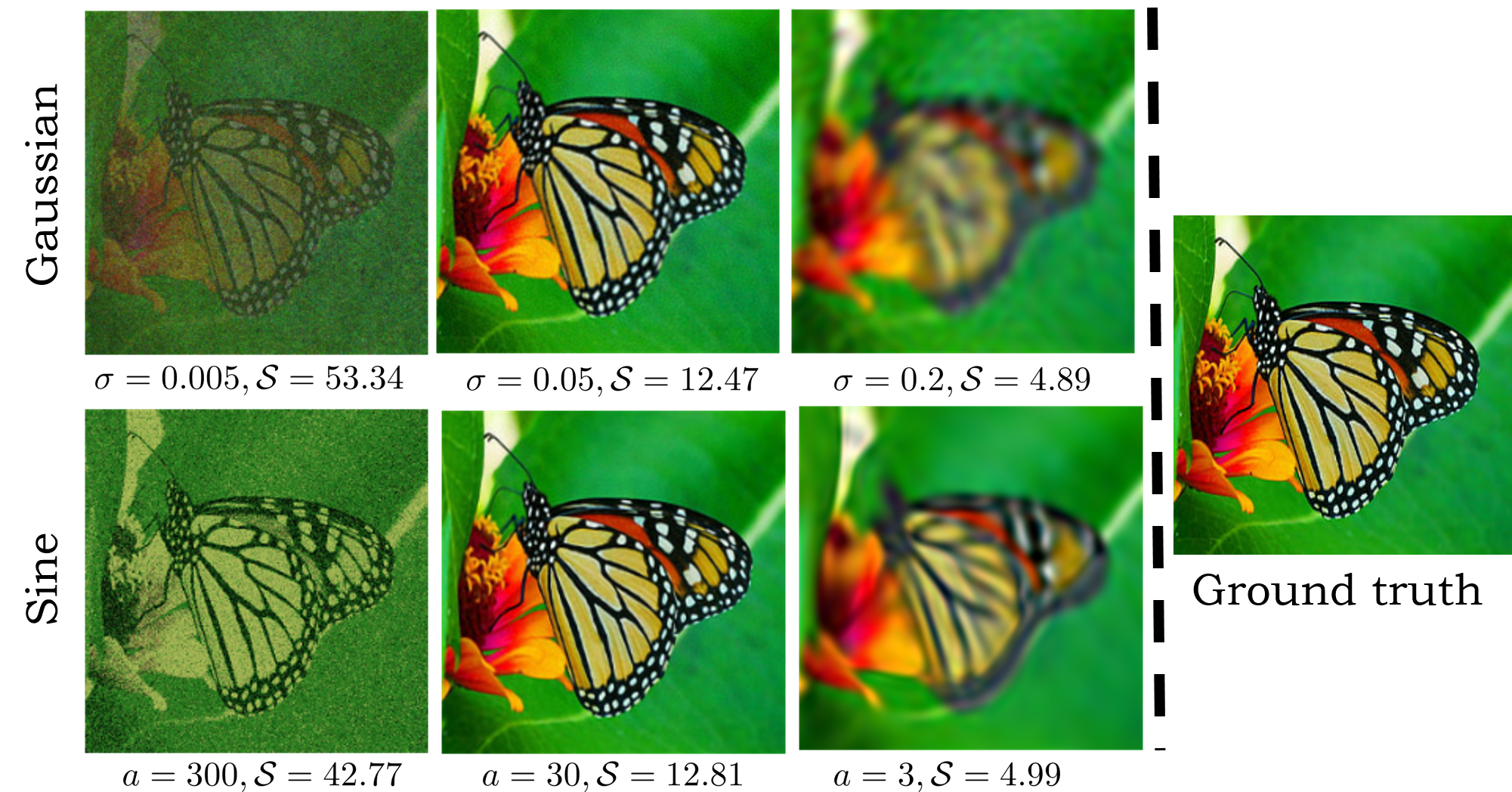}

    \caption{ \textbf{Stable rank ($\mathcal{S}$) vs the fidelity of reconstructions.} Having an extremely high or low stable rank (or equivalently a Lipschitz constant) hampers the ability of an MLP in encoding functions with fine details (Sec.~\ref{sec:Smoothness}). Thus, it is important to adjust the hyper-parameters of an activation function to tune the above metrics to a suitable range.}
    \label{fig:params}
\end{minipage}%
 \hfill
\begin{minipage}[t]{\dimexpr 0.5\textwidth-0.4\columnsep}
\centering
 \includegraphics[width=\columnwidth]{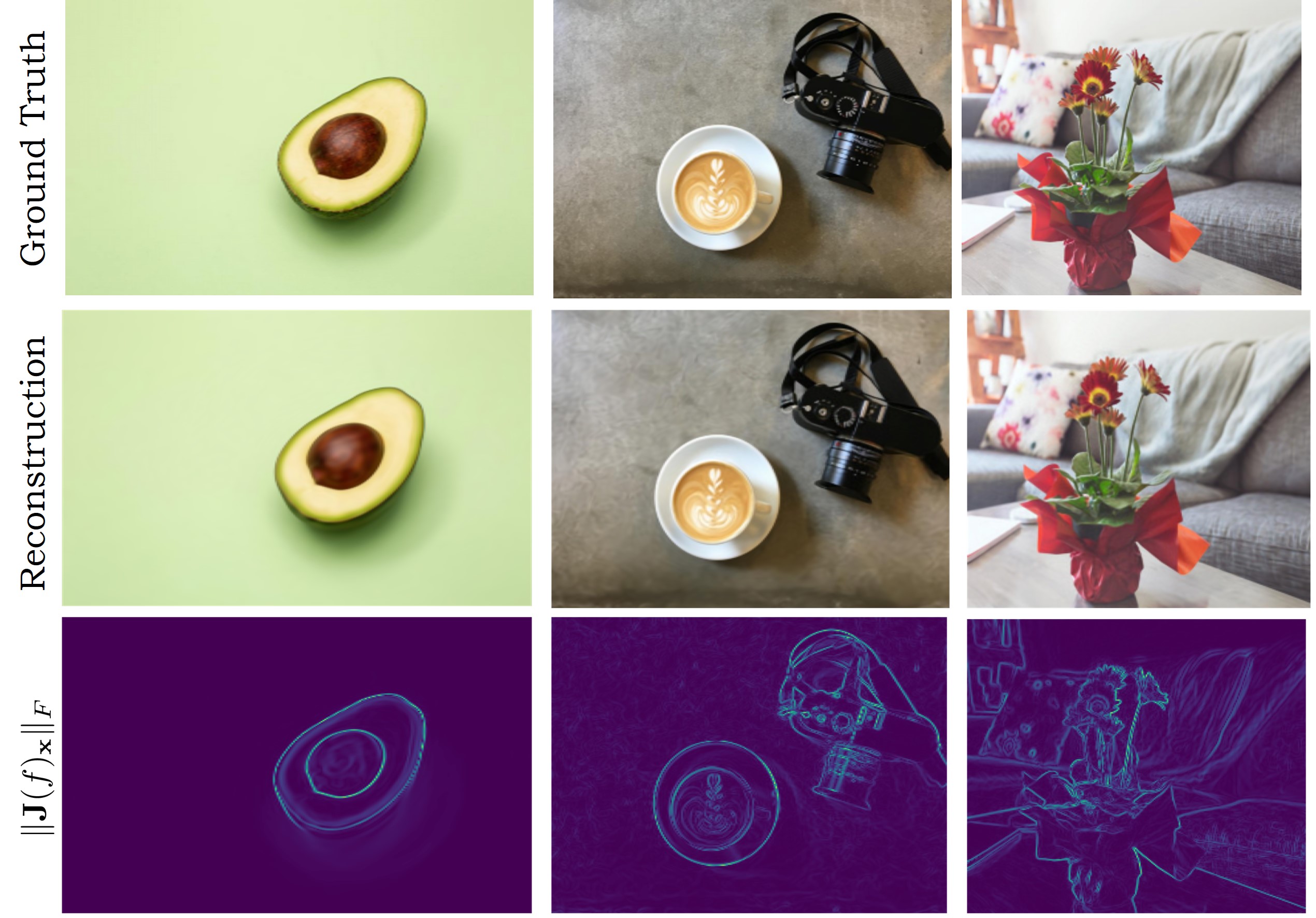}

    \caption{ \textbf{Distribution of the upperbound of the point-wise Lipschitz constant $\|\textbf{J}(f)_\x\|_F$  with Gaussian  reconstructions.} Having an activation function with a suitable bound for the local Lipschitz constant helps the network to learn functions with properly distributed derivatives.}
    \label{fig:grads}
\end{minipage}
\end{figure}

    

\subsection{Convergence}

\begin{figure}[!htp]
    \centering
    \subfloat{{\includegraphics[width=0.3\columnwidth]{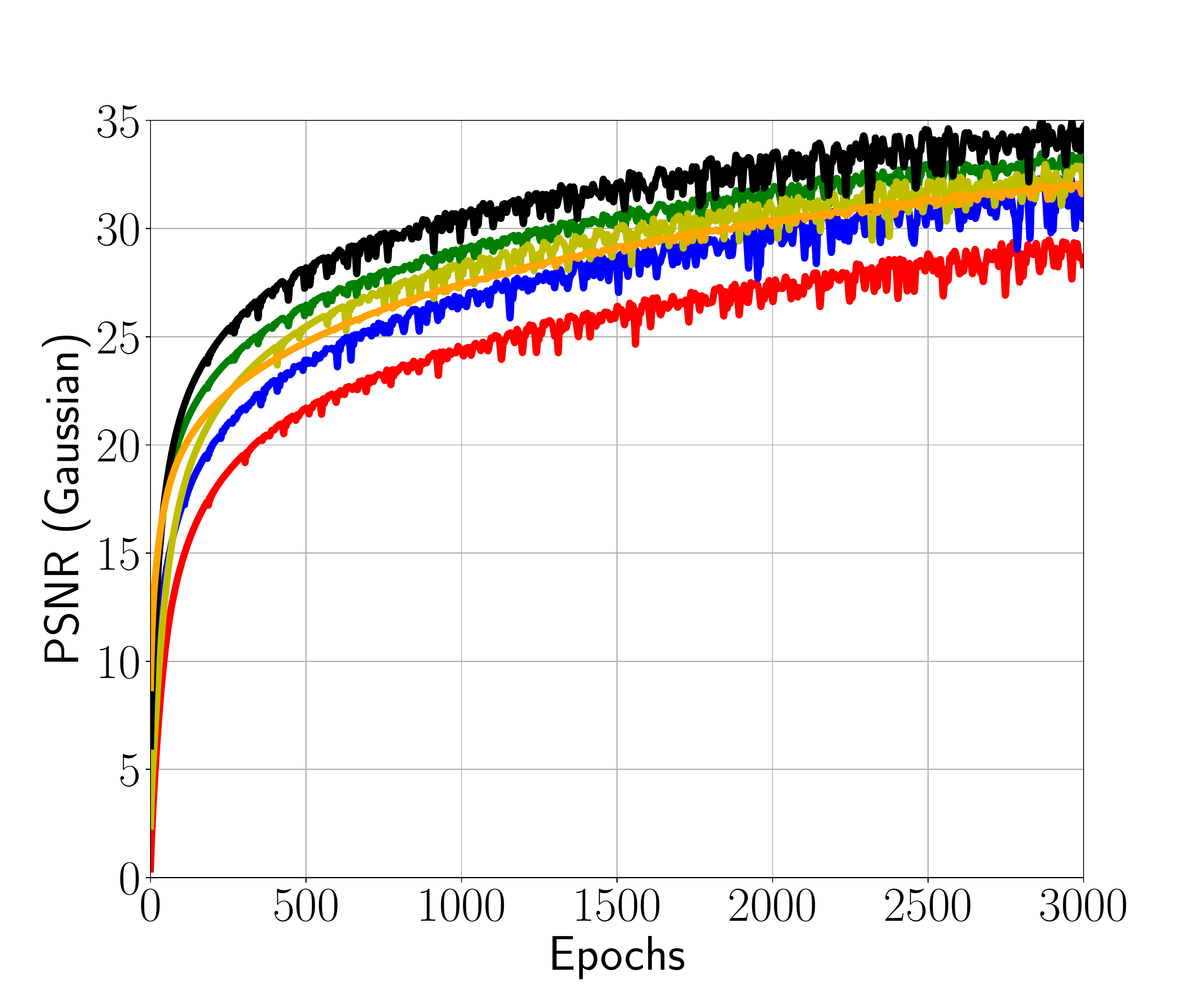} }}%
    \subfloat{{\includegraphics[width=0.3\columnwidth]{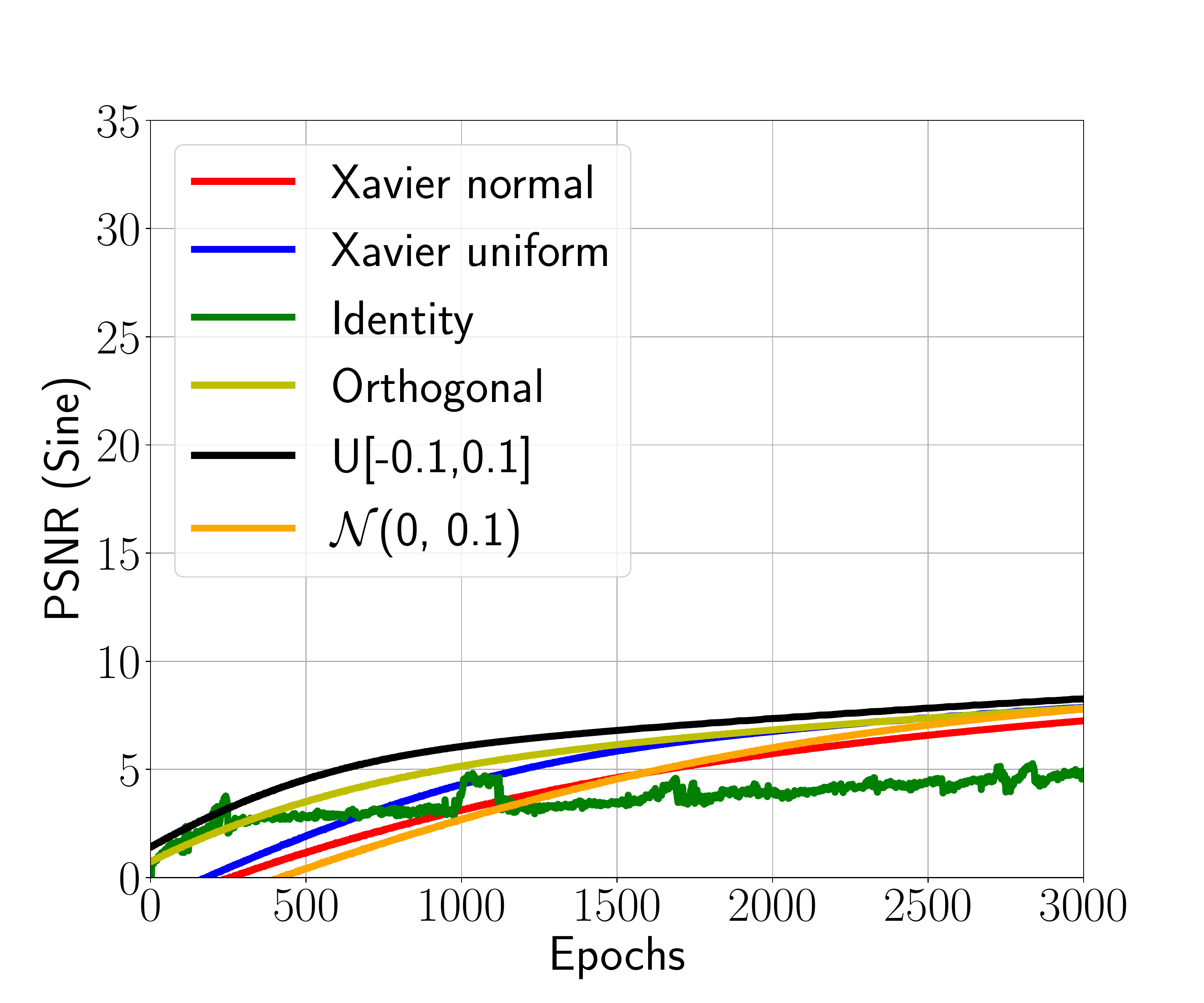} }}%
    \caption{ \textbf{Convergence rates of Gaussian and sine MLPs on natural images by  \cite{tancik2020fourier} under different initialization schemes.} Gaussian activations are significantly robust to various initialization methods compared to sine activations. Other proposed non-periodic activation functions (not shown in the figure) also demonstrate similar robustness. However, sine activations show a similar convergence rate to Gaussians when the MLPs are initialized strictly following the guidelines of \cite{sitzmann2020implicit}.}%
    \label{fig:convergencechart}
\end{figure}

Sitzmann \textit{et al.} comprehensively demonstrated that sine activations enable MLPs to encode signals with fine details. However, a drawback entailed with the sine activations is that they are extremely sensitive to the initialization of the MLP. In comparison, the proposed non-periodic activation functions do not suffer from such a problem. Fig.~\ref{fig:convergence} illustrates a qualitative example. When the initialization method of the MLP does not strictly follow the method proposed in Sitzmann \textit{et al.}, the sine activated MLPs do not converge even after $3000$ epochs. In contrast, Gaussian activations demonstrate much faster convergence. Fig.~\ref{fig:convergencechart} illustrates a quantitative comparison of convergence. We trained the networks on the natural image dataset released by \cite{tancik2020fourier} and the average PSNR value after each iteration is shown in Fig.~\ref{fig:convergencechart}. As clearly evident, the Gaussian activations enjoy higher robustness against the various  initialization schemes of an MLP.

\subsection{Local Lipschitz smoothness}

The local Lipschitz smoothness of a function converges to the Jacobian norm at the corresponding point (see Appendix). In Section \ref{sec:locallipschitz}, we showed that a good proxy measure for the Lipschitz constant is the range of the first-order derivative of the activation function. We further affirmed that the Lipschitz constant should be suitably chosen for better performance \textit{i.e.} a too high or too low Lipschitz constant can prevent the network from properly learning a signal.  Fig.~\ref{fig:params} illustrates an example that confirms this statement. When $\sigma$ increases,  $\mathrm{Range}|\psi'|$ of the Gaussian activation decreases, decreasing the Lipschitz constant (see Sec.~\ref{sec:locallipschitz}). In contrast, when $a$ increases, the $\mathrm{Range}|\psi'|$ of the sine activation increases, increasing the Lipschitz constant. A lower Lipschitz constant results in blurry edges as it does not allow sharp changes locally. On the other hand, an extremely large Lipschitz constant allows unwanted fluctuations. Hence, choosing the parameters to be in a suitable range is vital for better performance. Fig.~\ref{fig:grads} shows the distribution of local Lipschitz constants after encoding a signal with Gaussian activations with properly chosen parameters.

\section{Conclusion}

We seek to extend the current understanding of activation functions that allow coordinate-MLPs to encode functions with high fidelity. We show that the previously proposed sinusoid activation  \cite{sitzmann2019scene} is a single example of a much broader class of activation functions that enable coordinate-MLPs to encode high-frequency signals. Further, we develop generic guidelines to devise and tune an activation function for coordinate-MLPs and propose several non-periodic activation functions as examples. The proposed activation functions allow positional-embedding-free coordinate-MLPs, and show much better convergence properties against various initialization schemes compared to sinusoid activations. Finally, choosing Gaussian activations from the proposed list, we demonstrate compelling results across various signal encoding tasks.

\bibliographystyle{splncs04}
\bibliography{main}







\newpage
\appendix
\onecolumn
\section*{\Large Appendix}
\section{Encoding signals}

\begin{table}[!htp]
\centering
       \begin{tabular}{||c|c|c||}
\hline
Embedding & Image type & PSNR \\
\hline
\hline
ReLU &	Natural & 20.42  \\
Tanh & Natural & 16.91 \\
SoftPlus & Natural & 16.03 \\
SiLU & Natural & 17.59\\
\hline
Gaussian & Natural & 33.43 \\
Laplacian & Natural & 33.01  \\
Quadratic & Natural & 32.90   \\
Multi-Quadratic & Natural & 33.11 \\
ExpSin & Natural & 32.99 \\
Super-Gaussian & Natural & 33.12 \\
\hline
\hline
ReLU & Text & 18.49\\
Tanh & Text & 16.19 \\
SoftPlus & Text & 15.77 \\
SiLU & Text & 17.43 \\
\hline
Gaussian & Text & 36.17\\
Laplacian & Text & 36.29 \\
Quadratic & Text & 35.55\\
Multi-Quadratic & Text & 36.20\\
ExpSin & Text & 35.89 \\
Super-Gaussian & Text & 35.57\\
\hline
\hline
ReLU & Noise & 10.82 \\
Tanh & Noise & 9.66 \\
SoftPlus & Noise & 9.71 \\
SiLU & Noise & 11.21 \\
\hline
Gaussian & Noise & 11.78\\
Laplacian & Noise & 11.01\\
Quadratic & Noise & 11.67  \\
Multi-Quadratic & Noise & 11.29 \\
ExpSin & Noise &  11.45\\
Super-Gaussian & Noise & 11.33\\
\hline
\end{tabular}
  \caption{\textbf{Quantitative comparison  between activations in 2D signal encoding on \cite{tancik2020fourier} after 3000 epochs}. The proposed activations yield high PSNRs. The noise signals are difficult to be encoded with high fidelity due to limited redundancy.}
\label{tab:2dapp}
\end{table}

\begin{figure*}
    \centering
    \includegraphics[width = 1.\columnwidth]{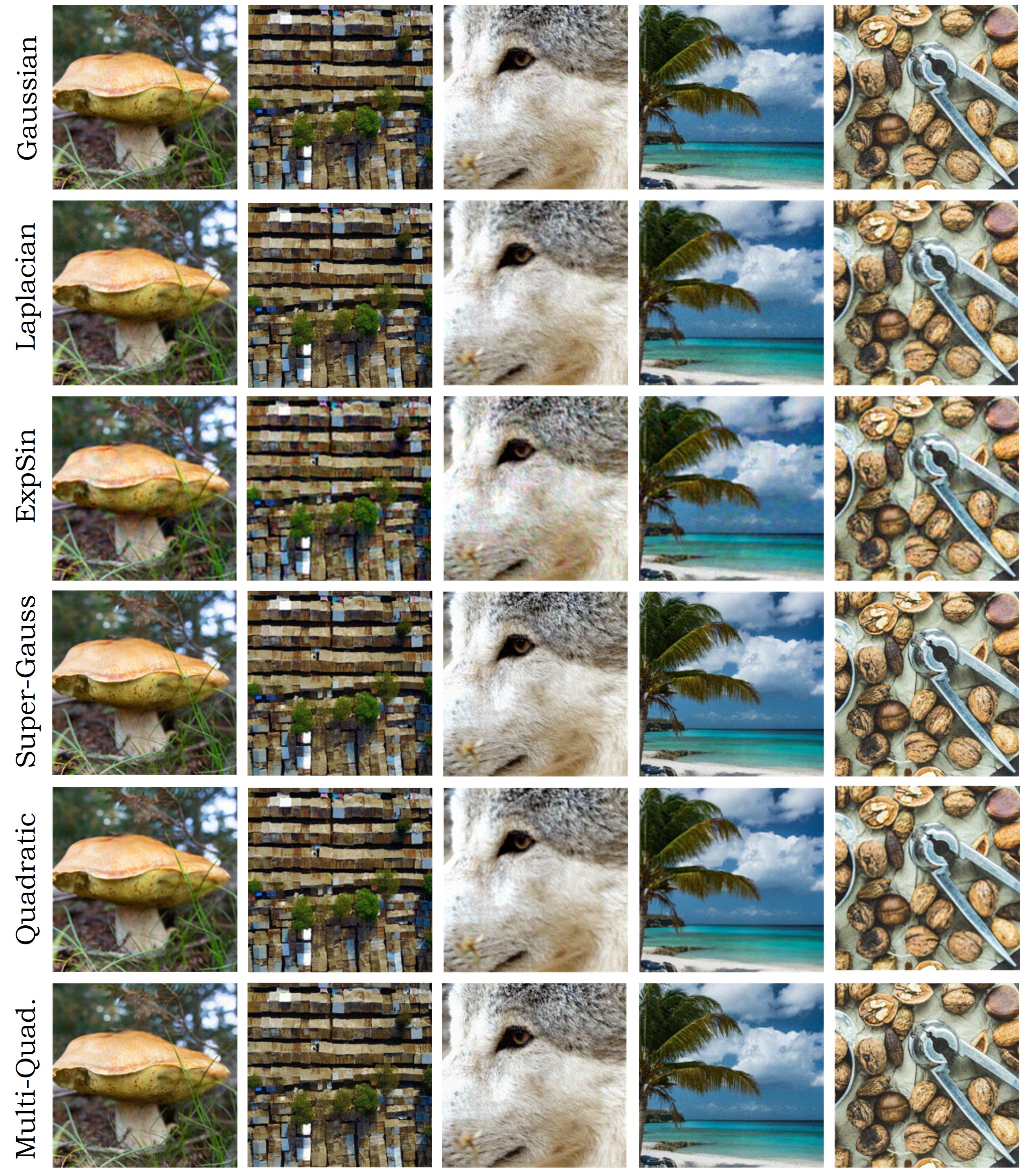}
    \caption{Qualitative examples of 2D signal encoding using the proposed activations on the natural images by \cite{tancik2020fourier}.}
    \label{fig:my_label}
\end{figure*}

 \begin{figure*}
    \centering
    \includegraphics[width = 1.\columnwidth]{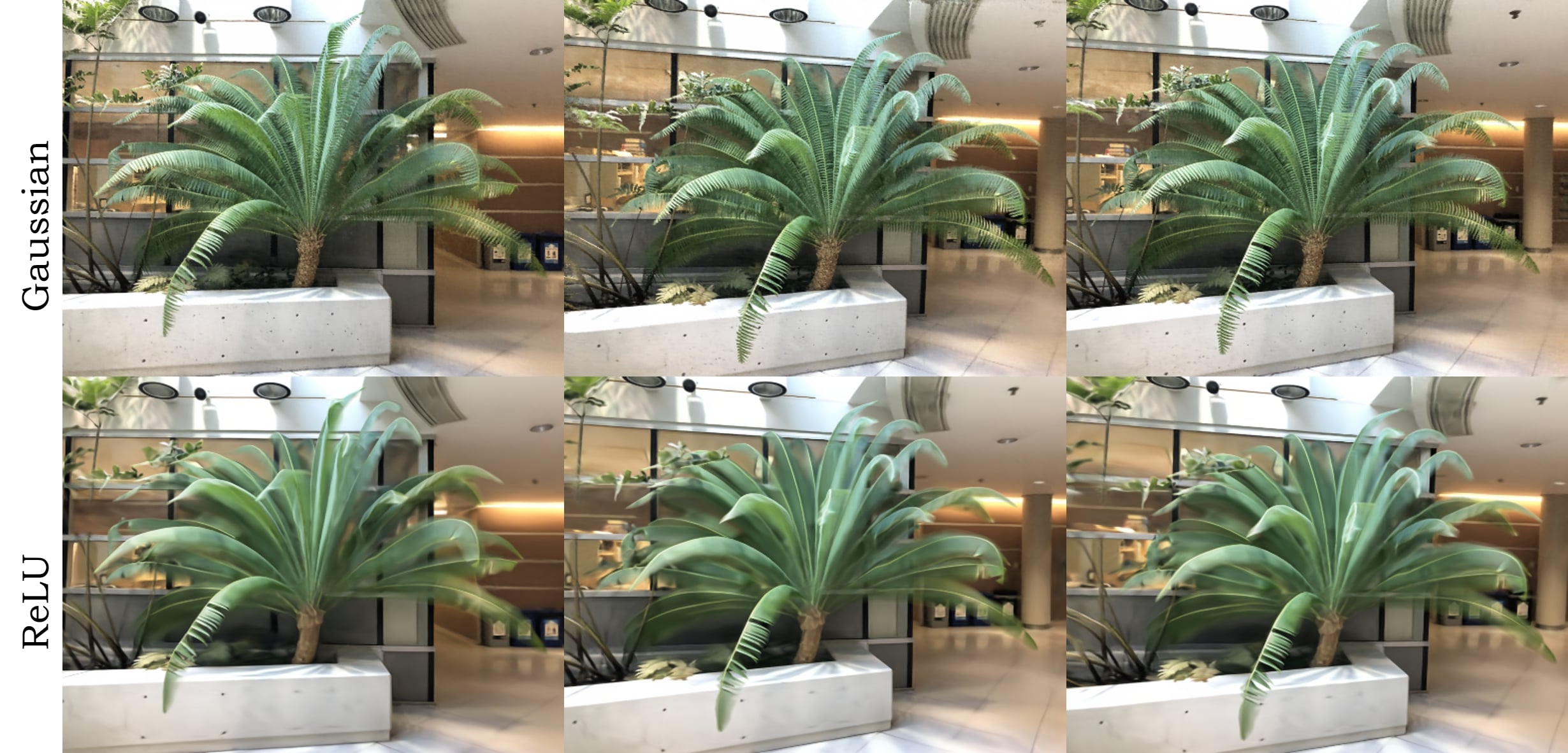}
    \caption{A qualitative comparison between ReLU and Gaussian activations (w/o positional embedding) in 3D view synthesis.}
    \label{fig:fern}
\end{figure*}

 \begin{figure*}
    \centering
    \includegraphics[width = 1.\columnwidth]{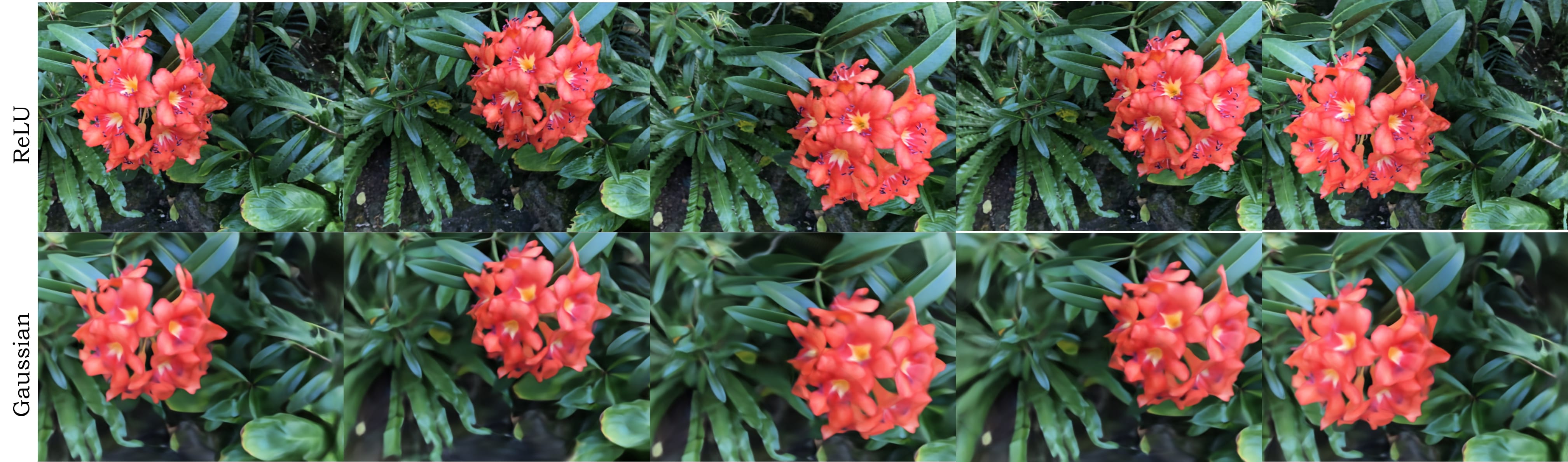}
    \caption{A qualitative comparison between ReLU and Gaussian activations (w/o positional embedding) in 3D view synthesis.}
    \label{fig:flower}
\end{figure*}

\begin{figure*}
    \centering
    \includegraphics[width = 1.\columnwidth]{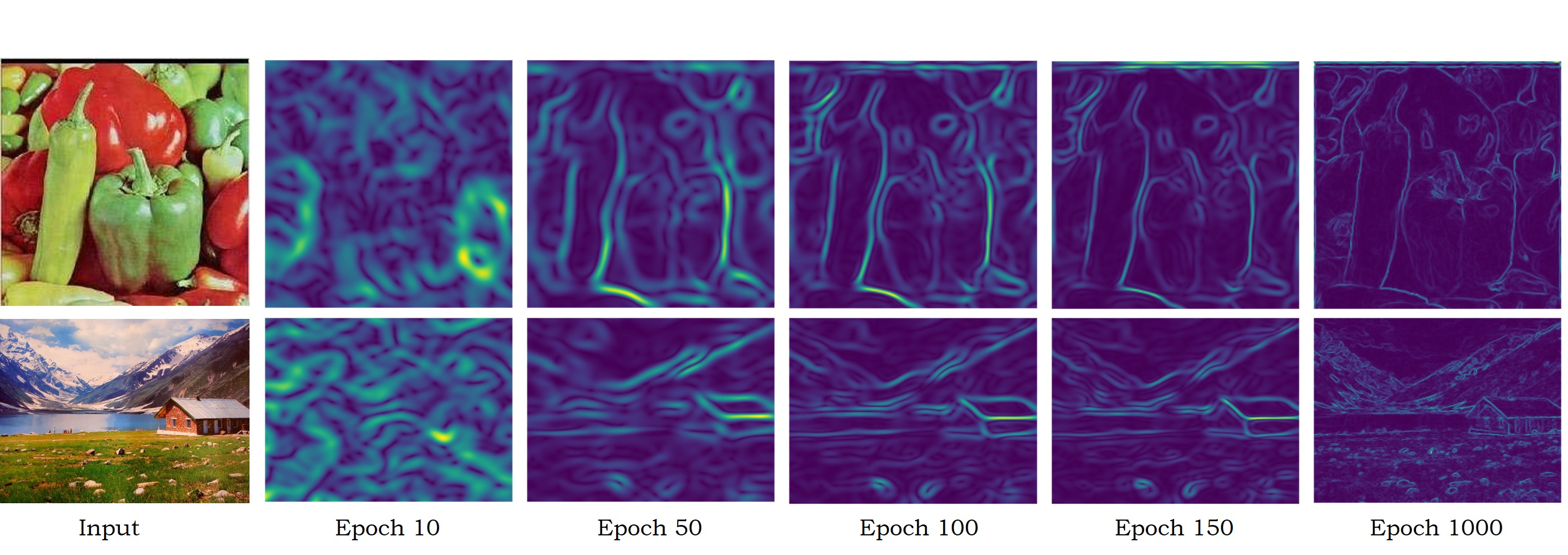}
    \caption{$\| \textbf{J}_f(\x) \|_F$ convergence as the training progresses.}
    \label{fig:my_label}
\end{figure*}

\begin{figure*}
    \centering
    \includegraphics[width = 1.\columnwidth]{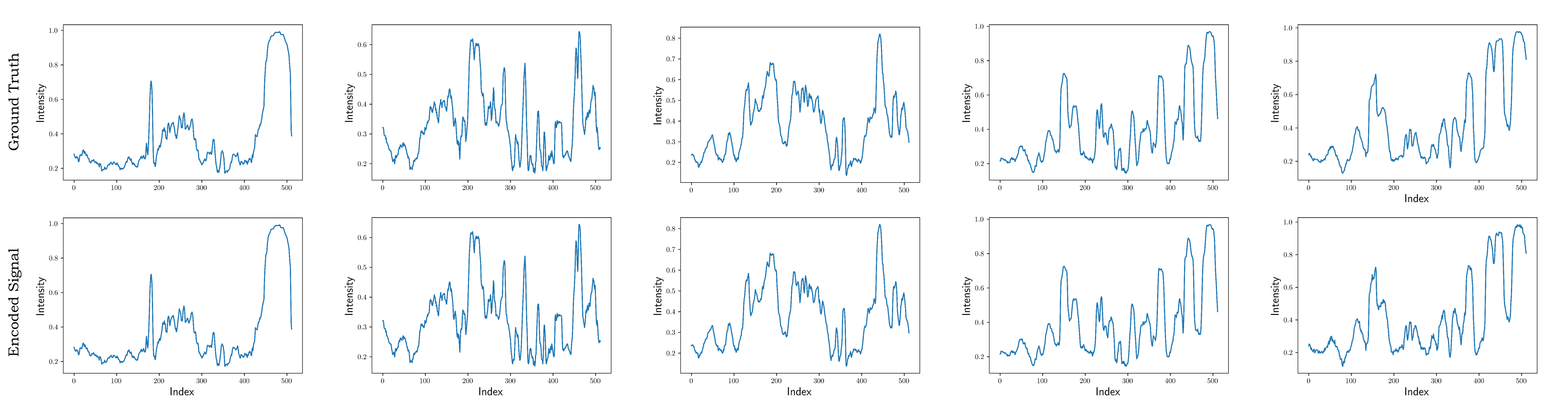}
    \caption{1D signal encoding using Gaussian activated MLPs.}
    \label{fig:my_label}
\end{figure*}

 \section{Norms of the layer outputs.}
 
\begin{figure*}
    \centering
    \includegraphics[width = 1.\columnwidth]{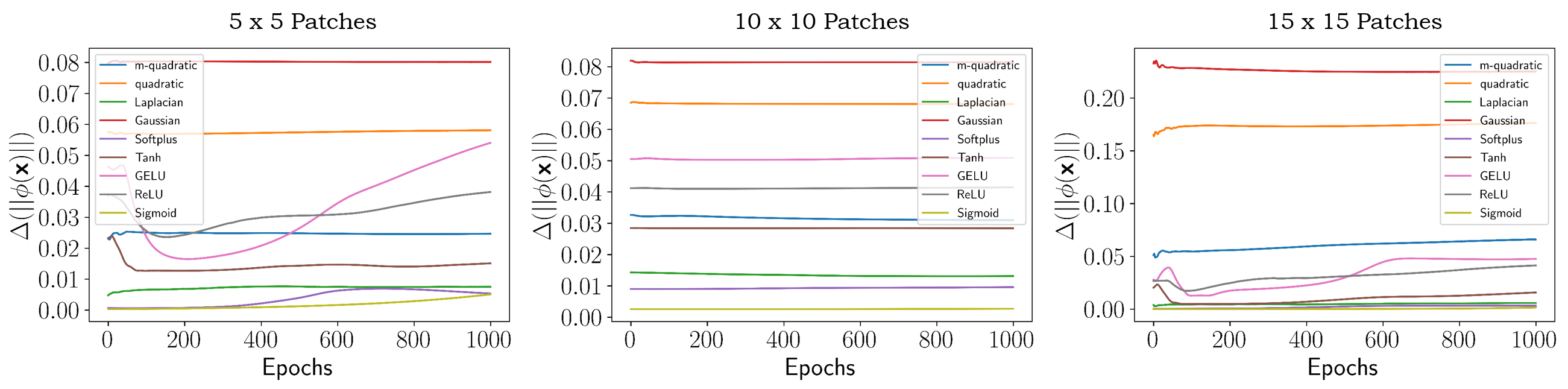}
    \caption{\textbf{The norms of the layer outputs remain approximately locally constant}. For each patch obtained via an overlapping sliding window over an image, we measure $\frac{\|\mathrm{max}(\phi(\x)) - \mathrm{min}(\phi(\x))  \|}{\|\mathrm{max}(\phi(\x)) \|}$. The above measure is then averaged over all the patches, layers, and a subset of $10$ images to obtain $\Delta(\| \phi(\x)\|)$ while training. As depicted, the norms of the vectors do not significantly deviate from their maximum within a subset.}
    \label{fig:epsilon}
\end{figure*}

 Our empirical results strongly suggested that the coordinate networks control the local Lipschitz constant of the encoded signal primarily via the angle between the layer outputs. In this section, we conduct another experiment to validate this behavior further. We measure the deviations of the layer output norms within patches of images (Fig.~\ref{fig:epsilon}). As shown, within a local area, the norms of the layer outputs do not significantly deviate from their maximum, which backs up our previous experiments.
 
In Sec~\ref{sec:locallipschitz}, the desired properties of the activations are derived under the restriction that the norms of the outputs of the layers remain approximately constant over a local subset. However, even in the case where the vector norms are not approximately constant, our conclusions still hold: First, let us formally define the local Lipschitz smoothness.

\begin{definition}
A function $f: \RB^{m} \to \RB^{n}$ is locally Lipschitz around $\x_0 \in \RB^m$ if for all $\x \in \mathbb{B}^{m}_{\delta}$ there exists a constant $C$ such that $\|f(\x) - f(\x_0)\| \leq C \|\x - \x_0\|$ where $\x_0$ is the center $\x_0$ of $\mathbb{B}^{m}_{\delta}$. Then, the smallest $C$ for which the above inequality is satisfied is called the Lipschitz constant of $f$ around $\x_0$, and is denoted as $C_{\x_0, \delta}(f)$.
\end{definition}

A hidden-layer is a composition of an affine function $g$ and a non-linearity $\psi$. Hence, the composite local Lipschitz constant of a hidden-layer is upper-bounded with $C_{\x_0, \delta}(\psi \circ g)  \leq C_{\x_0, \delta}(g)C_{\x_0, \delta}(\psi)$. We study these properties as $\delta$ approaches zero. As shown in Sec.\ref{sec:locallipschitz}, the affine transformation $g$ cannot vary the local Lipschitz smoothness across the signal. Thus, we focus on the activation function $\psi$. Since $\psi$ is a continuously differentiable function, applying the Taylor expansion gives

\begin{equation}
\label{eq:taylor}
    \psi(\x) = \psi(\x_0) + \textbf{J}(\psi)_{\x_0}(\x - \x_0) + \Theta(\|\x - \x_0\|),
\end{equation}

where $ \Theta(\|\x - \x_0\|)$ is a rapidly decaying function as $\x \to \x_0$. Rearranging Eq.~\ref{eq:taylor} we get

\begin{equation}
   \| \psi(\x) - \psi(\x_0) \| \leq \|\textbf{J}(\psi)_{\x_0}\|_o \|(\x - \x_0)\| + \|  \Theta(\|\x - \x_0\|) \|,
\end{equation}

\begin{equation}
\label{eq:lipdef}
  \underset{\delta \to 0}{\lim} \Bigg[  \underset{\x \in \mathbb{B}^m_{\delta}}{\sup}   \frac{\| \psi(\x) - \psi(\x_0) \|}{\| \x - \x_0 \|} \Bigg] \leq  \underset{\delta \to 0}{\lim} \Bigg[ \|\textbf{J}(\psi)_{\x_0}\|_o +  \frac{\|  \Theta(|\x - \x_0|) \|}{\| \x - \x_0 \|}  \Bigg].
\end{equation}

The left-hand side of Eq.~\ref{eq:lipdef} is the point-wise Lipschitz constant of $\psi(\cdot)$ at $\x_0$ by definition. Again, the quantity $\underset{\delta \to 0}{\lim}  \frac{\|  \Theta(|\x - \x_0|) \|}{\| \x - \x_0 \|}$ is zero by deifinition. Thus, denoting the point-wise Lipschitz constant of $\psi(\cdot)$ at $\x_0$ as $C_{\x_0}(\psi)$,  we get

\begin{equation}
\label{eq:op_inequal}
     C_{\x_0}(\psi) \leq \|\textbf{J}(\psi)_{\x_0}\|_o .
\end{equation}

Consider,

\[
   \| \textbf{J}(\psi)_{\x_0} \|_o = \underset{\mathbb{B}^m_{\delta} \ni \|\x\| = 1  }{\sup} \| \textbf{J}(\psi) \x\|
\]

Here the operator norm is defined with the vector-norm on the vector space of $\x$. Further, the Frobenius norm and the vector-norm are the same for vectors. Therefore,

\[
   \| \textbf{J}(\psi)_{\x_0} \|_o = \underset{\mathbb{B}^m_{\delta} \ni \|\x\| = 1  }{\sup} \| \textbf{J}(\psi) \x\|_F
\]

\noindent Frobenius norm is submultiplicative. Thus,

\begin{align*}
     \| \textbf{J}(\psi)_{\x_0} \|_o & \leq \underset{\mathbb{B}^m_{\delta} \ni \|\x\| = 1  }{\sup} \| \textbf{J}(\psi)\|_F \|\x\|_F \\
     & = \underset{\mathbb{B}^m_{\delta} \ni \|\x\| = 1  }{\sup} \| \textbf{J}(\psi)\|_F \|\x\| \\
     & = \| \textbf{J}(\psi)\|_F  \underset{\mathbb{B}^m_{\delta} \ni \|\x\| = 1  }{\sup} \|\x\| \\
     & =  \| \textbf{J}(\psi)\|_F. 
\end{align*}
   
Hence, with Eq.~\ref{eq:op_inequal}, we have 

\begin{equation}
\label{eq:llsa}
     C_{\x_0}(\psi) \leq \|\textbf{J}(\psi)_{\x_0}\|_F.
\end{equation}.

Since $\textbf{J}(\psi)$ is diagonal,

\begin{equation}
    C_{\x}(\psi) \leq \sqrt{D (\underset{x \in \RB}{\mathrm{max}}\psi'(x))^2},
\end{equation}

Where $D$ is the width of the network layer. One can see that in order to have a larger Lipschitz constant over some interval, the maximum first-order derivative of the activation should be higher over the same interval. Furthermore, in order to obtain varying local Lipschits smoothness, the maximum first-order derivatives of the activation within finite intervals should vary across the domain.

\end{document}